\documentclass[12pt]{iopart}
\usepackage{makeidx}
\usepackage{graphicx}
\usepackage{subfigure}
\usepackage{multirow}
\usepackage{color,soul} 
\usepackage{amssymb,amsfonts}
\usepackage{cases}
\usepackage{hyperref}
\usepackage{multirow}

\begin{document}

\title{RMSim: Controlled Respiratory Motion Simulation on Static Patient Scans}


\author{Donghoon Lee, Ellen Yorke, Masoud Zarepisheh, \\Saad Nadeem*, Yu-Chi Hu*}

\address{Department of Medical Physics, Memorial Sloan Kettering Cancer Center, New York, NY, USA}
\ead{\{leed10,yorkee,zarepism,nadeems,huj\}@mskcc.org}

\vspace{10pt}
\begin{indented}
\item[]*Corresponding Authors
\end{indented}

\begin{abstract}
\textbf{Objective:} This work aims to generate realistic anatomical deformations from static patient scans. Specifically, we present a method to generate these deformations/augmentations via deep learning driven respiratory motion simulation that provides the ground truth for validating deformable image registration (DIR) algorithms and driving more accurate deep learning based DIR.\\
\textbf{Approach:} We present a novel 3D Seq2Seq deep learning respiratory motion simulator (RMSim) that learns from 4D-CT images and predicts future breathing phases given a static CT image. The predicted respiratory patterns, represented by time-varying displacement vector fields (DVFs) at different breathing phases, are modulated through auxiliary inputs of 1D breathing traces so that a larger amplitude in the trace results in more significant predicted deformation. Stacked 3D-ConvLSTMs are used to capture the spatial-temporal respiration patterns. Training loss includes a smoothness loss in the DVF and mean-squared error between the predicted and ground truth phase images. A spatial transformer deforms the static CT with the predicted DVF to generate the predicted phase image. 10-phase 4D-CTs of 140 internal patients were used to train and test RMSim. The trained RMSim was then used to augment a public DIR challenge dataset for training VoxelMorph to show the effectiveness of RMSim-generated deformation augmentation. \\
\textbf{Main results:} We validated our RMSim output with both private and public benchmark datasets (healthy and cancer patients). The structure similarity index measure (SSIM) for predicted breathing phases and ground truth 4D CT images was 0.92$\pm$0.04, demonstrating RMSim's potential to generate realistic respiratory motion. Moreover, the landmark registration error in a public DIR dataset was improved from 8.12$\pm$5.78mm to 6.58mm$\pm$6.38mm using RMSim-augmented training data.\\ 
\textbf{Significance:} The proposed approach can be used for validating DIR algorithms as well as for patient-specific augmentations to improve deep learning DIR algorithms. The code, pretrained models, and augmented DIR validation datasets will be released at \url{https://github.com/nadeemlab/SeqX2Y}. The supplementary video can be found at \url{https://youtu.be/xIx8B_Q_R9o}.

\end{abstract}

%
%
%
%
%

\section{Introduction}
{Respiratory motion hampers accurate diagnosis as well as image-guided therapeutics. For example, during radiotherapy,} it may lead to poor local tumor control and increased radiation toxicity to the normal organs~\cite{motionlung2018}. 
It can also exhibit itself as motion artifacts in the acquired images, making it difficult to differentiate nodule/tumor morphology changes from those induced by respiratory motion. This also makes the image registration task across different breathing phases as well as across different time points challenging. To validate the image registration accuracy/performance for commissioning solutions available in clinical commercial systems, the American Association of Physicists in Medicine(AAPM) TG-132~\cite{Brock2017TG132} recommended independent quality checks using digital phantoms. Current commercial solutions such as ImSimQA allow creation of synthetic deformation vector fields (DVFs) by user-defined transformations with only a limited degree of freedom. These monotonic transformations can not capture the realistic respiratory motion.

For modeling respiration motion, an intuitive representation of motion is time-varying displacement vector fields (DVFs) obtained by deformable image registrations (DIR) in 4D images, acquired in a breathing cycle. Surrogate-driven approaches~\cite{MCCLELLAND201319} {employ} DVF as a function of the surrogate breathing signal. However, an exact and direct solution in the high-dimensional space of DVFs is computationally intractable. {Still, motion surrogates have been widely studied in the field of radiotherapy for building models establishing the relationship between surrogates and respiratory motion estimated from the image data \cite{MCCLELLAND201319}. For example, the 1D diaphragm displacement has been reported as a reliable surrogate for tumor motion model \cite{Cervi_o_2009} as well as for PCA (principle component analysis) respiratory motion model to correct CT motion artifacts~\cite{Zhang2007_PCA}.}

Recently, Romaguera et al.~\cite{ROMAGUERA2020_2DSeq2Seq} used a 2D {sequence-to-sequence (Seq2Seq) network~\cite{seq2seq2014}} to predict 2D in-plane motion for a single future time point. Krebs et al.~\cite{Krebs2020_cVAE} applied a similar encoder-decoder network in a conditional variational autoencoder (cVAE) framework {in which network parameters were learned to approximate the distribution of deformations in low-dimensional latent space with the encoder and decode the latent features} for {2D motion prediction with the decoder.} Romaguera et al.~\cite{ROMAGUERA2021_cVAE} integrated Voxelmorph \cite{balakrishnan2019voxelmorph} for assisting the VAE encoder to map deformations in latent space conditioned on anatomical features from 3D images. Temporal information of 2D surrogate cine images from a 2D Seq2Seq network was used to predict 3D DVF {at a single future time point.} 

{In this paper, we present a novel deep learning respiratory motion simulator (RMSim) that learns to generate realistic patient-specific respiratory motion represented by time-varying DVFs at different breathing phases from a static 3D CT image. For the first time,} we also allow modulation of this simulated motion via arbitrary 1D breathing traces as auxiliary input to create large variations. This in turn creates diverse patient-specific data augmentations while also generating ground truth for DIR validation. Our work has several differences and advantages over the aforementioned deep learning approaches: {(1) we used 3D Seq2Seq architecture for the first time which has never been attempted before for predicting deformations due to GPU memory limitations, (2) we did not use VoxelMorph in its entirety but only the Spatial Transform module to train our model end-to-end, and (3) as opposed to predicting just a single future time point, we can predict 9 future time point breathing phases simultaneously (learnt from 4D-CT images with 10 3D CT breathing phases) along with their 3D DVFs. We have thoroughly  validated our RMSim output with both private and public benchmark datasets (healthy and cancer patients) and demonstrated that adding our patient-specific augmentations to training data can improve performance/accuracy of state-of-the-art deep learning DIR algorithms. We also showcase breathing trace-modulated respiratory motion simulations for public static radiology scans (in the accompanying \textbf{supplementary video}). The code, pretrained models, and augmented DIR validation datasets will be released at 
\url{https://github.com/nadeemlab/SeqX2Y}.


\begin{figure}[th!]
\begin{center}
\footnotesize
\setlength{\tabcolsep}{3pt}
\includegraphics[width=1\textwidth]{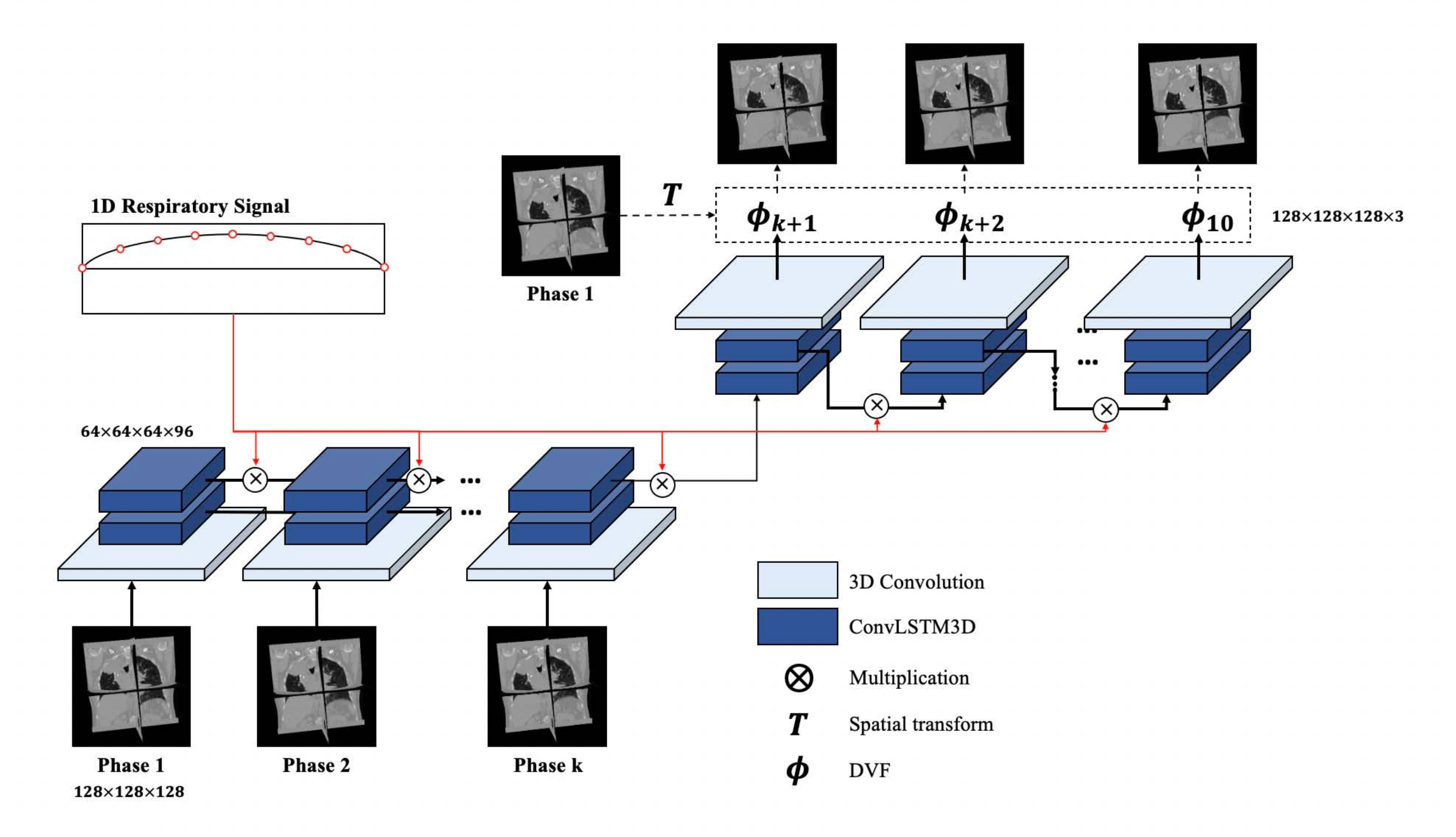}
\caption{The schematic image for the proposed deep learning model. The Seq2Seq encoder-decoder framework was used as the backbone of the proposed model.  The model was built with 3D convolution layers {for feature encoding and output decoding} and 3D convolutional Long Short-Term Memory (3D ConvLSTM) layers {for spatial-temporal correlation between time points}. The last layer of the decoder was a spatial transform layer to warp the initial phase image with the predicted Deformation Vector Field (DVF). To modulate the respiratory motions the 1D breathing trace was given as input along with the initial phase image. {The dimension of image volume was 128 $\times$ 128 $\times$ 128 and the input feature to 3D ConvLSTM is 64 $\times$ 64 $\times$ 64 $\times$ 96 (Depth $\times$ Width $\times$ Height $\times$ Channel)} }
\label{fig:Model}
\end{center}
\end{figure}

\section{Materials and Methods}
\subsection{Datasets}
We used an internal lung 4D-CT dataset retrospectively collected and de-identified from 140 non-small cell lung cancer (NSCLC)  patients receiving radiotherapy in our institution. The {helical and cine mode} 4D-CTs were acquired using Philips Brilliance Big Bore or GE Advantage {respectively} and binned into 10 phases using the vendor's proprietary software with breathing signals from bellows or external fiducial markers. The x-ray energy for the CT image was 120 kVp {and tube current varies case by case according to vendor-specific tube current modulations based on patient size. The mAs range is [100, 400] for GE and [500, 800] for Philips.} The image slice dimension was 512x512, while the number of image slices varied patient by patient. We used the 100:40 split for training:testing. 

We used 20 cases of the Lung Nodule Analysis (LUNA) challenge dataset~\cite{SETIO20171LUNA} {containing 3D radiology CTs for lung tumor screening} to show that our RMSim model trained with the internal dataset can be effectively applied to an external radiology/diagnostic dataset to generate realistic respiration motions (see accompanying \textbf{supplementary video}). {For quantitative evaluation of the model generality on an external data set, we used POPI \cite{vandemeulebroucke2011spatiotemporal} dataset which contains 6 10-phase 4D-CTs with segmented lung masks as well as annotated landmarks on the vessel and airway bifurcations.} 
 
{To validate} the effectiveness of data augmentation using synthetic respiratory motion images generated from our RMSim model in the deformable registration task, we used the Learn2Reg 2020 challenge dataset~\cite{hering_alessa_2020_Learn2Reg}. The Learn2Reg dataset consists of 30 subjects (20 for the training / 10 for the testing) with 3D CT thorax images taken in inhale and exhale phases. For each Learn2Reg 20 inhale/exhale pairs, we generated other phases of images using our RMSim model which was trained with the internal dataset, therefore increasing the sample size to 200 in total to augment the training of a well-known unsupervised deep learning {DIR} method, VoxelMorph~\cite{balakrishnan2019voxelmorph}. Unfortunately the inhale-exhale landmarks are not publicly available {in Learn2Reg dataset to assess the registration accuracy}. For the landmarks evaluation {in registration task}, we used the POPI dataset. {Brief description/purpose of all the datasets used in this study is given in Table~\ref{table:datasets}.} All datasets used in this study were cropped to eliminate the background and resampled to 128$\times$128$\times$128 with 2mm voxel size {due to the GPU memory constrains}.

\begin{table*}[ht]
\centering
\caption{{Datasets used in this study.}}
\label{table:datasets}
\footnotesize
\begin{tabular}{l|p{0.15\linewidth}|p{0.20\linewidth}|p{0.25\linewidth}|p{0.15\linewidth}}
\hline
\textbf{Dataset} &  \textbf{Size} & \textbf{Description} & \textbf{Purpose} & \textbf{Evaluation} \\
\hline
Internal 4D-CTs  & 140 (100 training, 40 testing) & 10-phase radiotherapy 4D-CTs & Training and testing RMSim & Image similarity\\
\hline
LUNA & 20 & Radiology CTs for lung nodule detection & Testing model generality & Visualization and qualitative \\
\hline
POPI 4D-CTs & 6 & 10-phase 4D-CTs with landmarks & Testing model generality (evaluating DVF accuracy) & Target Registration Error (TRE) of landmarks \\
\hline
Learn2Reg  & 30 (20 training, 10 testing) & Inspiration-expiration thorax CT pairs with lung segmentations & Training and testing RMSim-augmented deep learning deformable image registration  (Voxelmorph) & Lung segmentation (Dice score) and image similarity \\
\hline
\end{tabular}
\end{table*}

\subsection{Realistic Respiratory Motion Simulation}

\begin{figure}[t!]
\begin{center}
\footnotesize
\setlength{\tabcolsep}{3pt}
\includegraphics[width=1\textwidth]{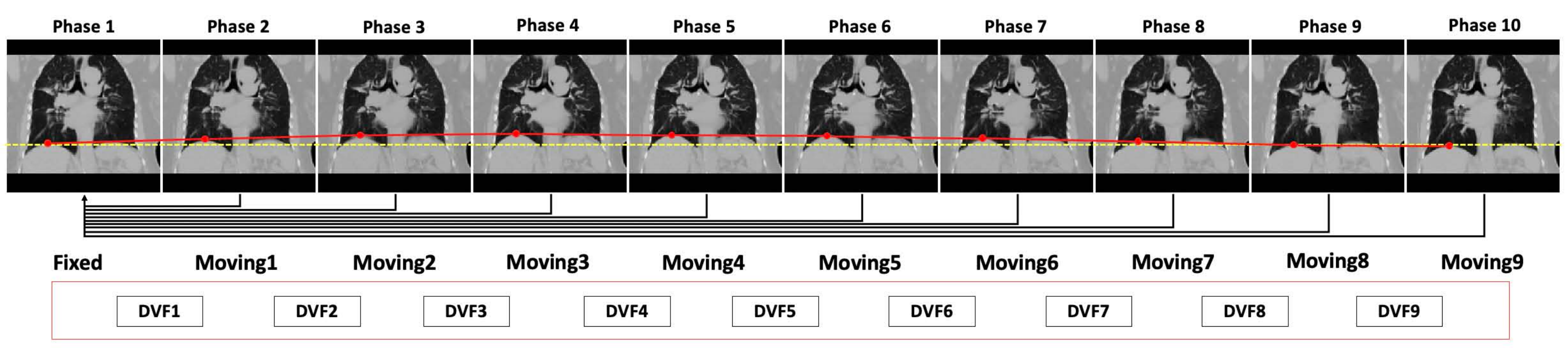}
\caption{Respiration motion surrogate extraction using a diaphragm point that has the maximum superior-inferior displacement across the phases. LDDMM was used to register the phase 1 (fixed) image to other phases (moving) to get the DVFs. The diaphragm point's trajectory in z-axis (shown in red) across the phases was considered as the breathing trace. Yellow line shows the diaphragm position at phase 1.}
\label{fig:RPM}
\end{center}
\end{figure}

{ Sequence-to-Sequence (Seq2Seg) is a many-to-many network architecture originally developed for natural language processing tasks such as language translation. Inspired by Seq2Seq,  the proposed RMSim, illustrated in Figure~\ref{fig:Model}, is a novel deep learning encoder-decoder architecture that comprises three main parts including 3D convolution, ConvLSTM3D (3D Convolutional Long-Short Term Memory), and spatial transformation layer (adapted from VoxelMorph \cite{balakrishnan2019voxelmorph}). The 3D convolution in the encoder is used to reduce the matrix dimension and extract salient features from images. We used 3$\times$3$\times$3 kernel size and 2$\times$2$\times$2 stride size to reduce the matrix dimension to 1/8. The number of channels for 3D convolution layer is 96. LSTM has a more complex cell structure than a neuron in classical recurrent neural network (RNN). Apart from the cell state, it contains gate units to decide when to keep or override information in and out of memory cells to better handle the gradient vanishing problem in recurrent neural network. This helps in learning long term dependencies. ConvLSTM \cite{ShiCovLSTMNIPS2015} replaces Hadamard product with convolution operators in the input as well as the state transitions to capture the spatial pattern of the feature representations aggregated from different time points. We implemented ConvLSTM in 3D for handling the 3D phase images from the 4D-CT. 
We used two stacked ConvLSTM3D layers to make the network deeper, adding levels of abstraction to input observations similar to the typical deep neural network. The hidden state output from ConvLSTM3D was fed to both the next layer in the same stack and the next timepoint ConvLSTM3D layer. The output of ConvLSTM3D in the decoder at each predicted time point was up-sampled to the original input resolution and output channels were reduced via 3D convolution, resulting in the 3D DVF for the final output. The initial phase CT image was then deformed to a predicted phase image at different breathing phase using spatial transformation layer and the predicted 3D DVFs.}


Moreover, to modulate the predicted motion with a patient-specific pattern, we used an auxiliary input of 1D breathing trace. In this paper, we considered the amplitude of diaphragm apex motion as the surrogate of the respiratory signal~\cite{Cervi_o_2009}. {The 1D breathing trace for each training case was extracted using DVF obtained from} large deformation diffeomorphic metric mapping (LDDMM) DIR provided by ANTs (Advanced Normalization Tools). {Specifically, using the DVF, the apex point in diaphragm was propagated from the phase at the end of inhalation to other phases to generate the 1D displacement trace. The apex of the diaphragm was determined by finding the lung surface voxel with the maximum superior-inferior (z-axis) displacement among the DVFs. The z-axis displacement of the apex voxel at each phase resembles the 1D breathing trace.} Figure~\ref{fig:RPM} describes the process of preparing the 1D respiratory signal. {Feature-wise transformations, e.g. addition or multiplication, are simple and effective mechanisms to incorporate conditioning information from another data source to the features learned in the network.} In this paper, the hidden state of ConvLSTM at each phase is modulated by a simple element-wise multiplication of the phase-amplitude of the trace:
\begin{equation}
\label{eqn:modulation}
m(H_t,b_t) =b_{t}H_t,
\end{equation}
where $H_t$ is the hidden state encoded from the sequence of phase images up to phase $t$ and $b_t$ is the amplitude of the breathing trace at phase $t$,

The loss function for training includes the mean-squared error of ground truth phase image and predicted phase image, and the regularization on the gradient of DVF by promoting smoothness of DVF:
\begin{equation}
\label{eqn:loss}
Loss =\sum_{t>0}[(Y_t-T(X_0,\phi_t))^2 + ||\nabla\phi_t||^2],
\end{equation}
where $X_0$ is the initial phase image (phase 1 in this paper), $T$ is the spatial transform (adapted from VoxelMorph), $\phi_t$ is the predicted DVF for phase $t$ and $Y_t$ is the ground truth phase image at phase $t$.

We developed RMSim using the PyTorch library (version 1.2.0). We used Adam for optimization and set learning rate to be 0.001 (as done in the original Seq2Seq paper~\cite{ShiCovLSTMNIPS2015}). Due to the large data size of 4D image sequence (10 3D CT phase images constituting a single 4D-CT), the batch size was limited to 1 and the number of feature channels was 96, considering GPU memory and training time. The model was trained and tested on an internal high performance computing cluster with 4 NVIDIA A40 GPUs with 48GB memory each. Our model consumed 35.2 GB GPU memory and the training time was approximately 72 hours. The inference time for 9 phases and 40 total test cases from the internal dataset was less than 3 minutes. 

\subsection{Data augmentation by RMSim}
Since RMSim can generate a series of realistic respiratory motion-induced images from a single 3D CT, one of its use cases is data augmentation for training DIR algorithms. For each of the 20 training cases in the Learn2Reg Grand Challenge dataset~\cite{hering_alessa_2020_Learn2Reg}, we randomly selected a 1D breathing trace from our internal dataset to modulate the motion {on the Learn2Reg inhalation image} to generate 9 additional phase images, increasing the training size 10-fold. {We chose a popular deep learning DIR method, VoxelMorph, suitable for unsupervised training for the propose of validating effectiveness of data augmentation.}  We first trained a VoxelMorph model with the original 20 inhalation-to-exhalation image pairs in the Learn2Reg training set. We then trained another VoxelMorph model with the augmented data including 200 pairs of inhalation-to-phase images. 
We compared the registrations from the two VoxelMorph models for validating the effectiveness of data augmentation.

\subsection{Evaluation Metrics} 
For image similarity, we used structure similarity index measure (SSIM) \cite{SSIM2004} which measures the similarity of two given images based on the degradation of structural information, including luminance, contrast and structure. The closer the SSIM value is to 1, the more similarity between the two images. SSIM was used for comparing RMSim-predicted phase images and ground truth phase images in the internal test cases. SSIM was also used for comparing deformable registration results from VoxelMorph to validate data augmentation effectiveness in Learn2Reg test cases, which additionally were evaluated with the provided lung segmentation using Dice score to compare the ground truth lung contours and propagated lung contours.

For landmark comparison in the POPI dataset, we used Target Registration Error (TRE), defined as the Euclidean distance between a landmark
position spatially transformed and the target position.



\section{Results}
For each test case in the internal 4D-CT dataset, we generated 9 simulated phase images from the ground truth phase 1 image by deforming the phase 1 image using the predicted DVF at each phase. We calculated SSIM to measure the image similarity (SSIM\textsuperscript{sim}) between the simulated phase image and the ground truth phase image. For comparison, we also calculated the SSIM (SSIM\textsuperscript{gnd}) between the ground truth phase 1 image and the rest of the ground truth phase images. The average SSIM\textsuperscript{sim} was 0.92$\pm$0.04, compared to 0.86$\pm$0.08 of SSIM\textsuperscript{gnd} ($p <0.01$.) 

{We also measured the diaphragm displacement between the reference respiratory signal and the predicted signal (see Figure~\ref{fig:error}). As can be seen, the error increased from inhale to exhale phases. This is because prediction accuracy decreases at later time points. However, the overall displacement error was within 3 mm. Adding more realistic respiratory data for training can further reduce this displacement error.}

\begin{figure}[th!]
\begin{center}
\footnotesize
\setlength{\tabcolsep}{3pt}
\includegraphics[width=0.8\textwidth]{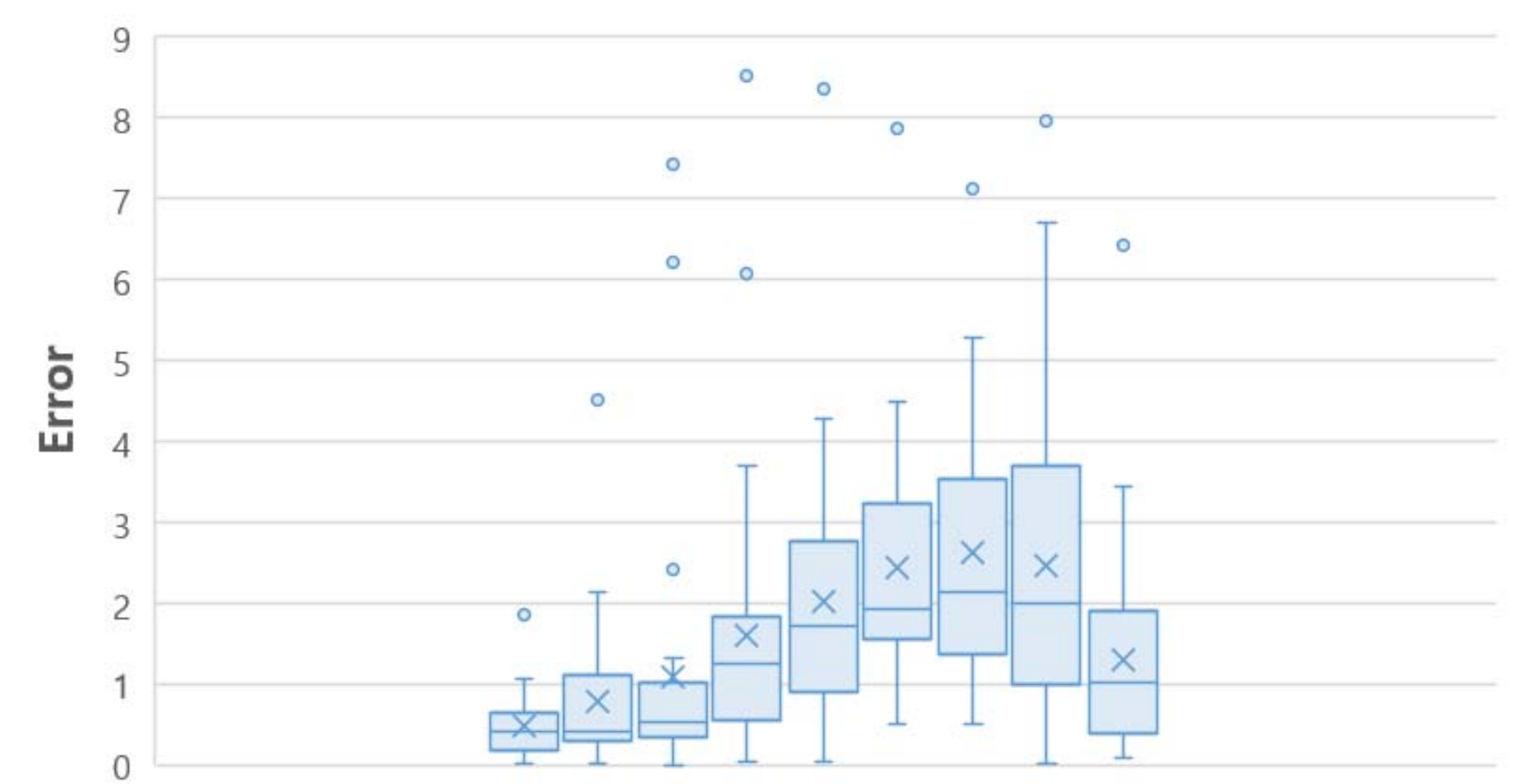}
\caption{{The error between reference respiratory signal (diaphragm displacement in mm) and predicted signal.}}
\label{fig:error}
\end{center}
\end{figure}

To demonstrate the modulation flexibility of the 1D breathing traces, we applied {different} breathing traces to the same 3D CT image to generate different motion simulations, as shown in Figure~\ref{fig:Results}. The plot on the top illustrates the two 1D breathing traces used for modulation. The breathing trace 1 (BT1), denoted by orange color line, represents the original respiratory signal for the case. BT2 denoted by gray line is a trace from another patient that was used to generate the simulated images. The white horizontal line indicates the position of the apex of the diaphragm in the initial phase (the first column). It is used as a reference to show the relative positions of the diaphragm at different phases. The diaphragm in images on the upper row clearly shows the more significant movement as BT2 has higher amplitudes in the trace.

\begin{figure}[!ht]
\begin{center}
\footnotesize
\setlength{\tabcolsep}{3pt}
\includegraphics[width=1\textwidth]{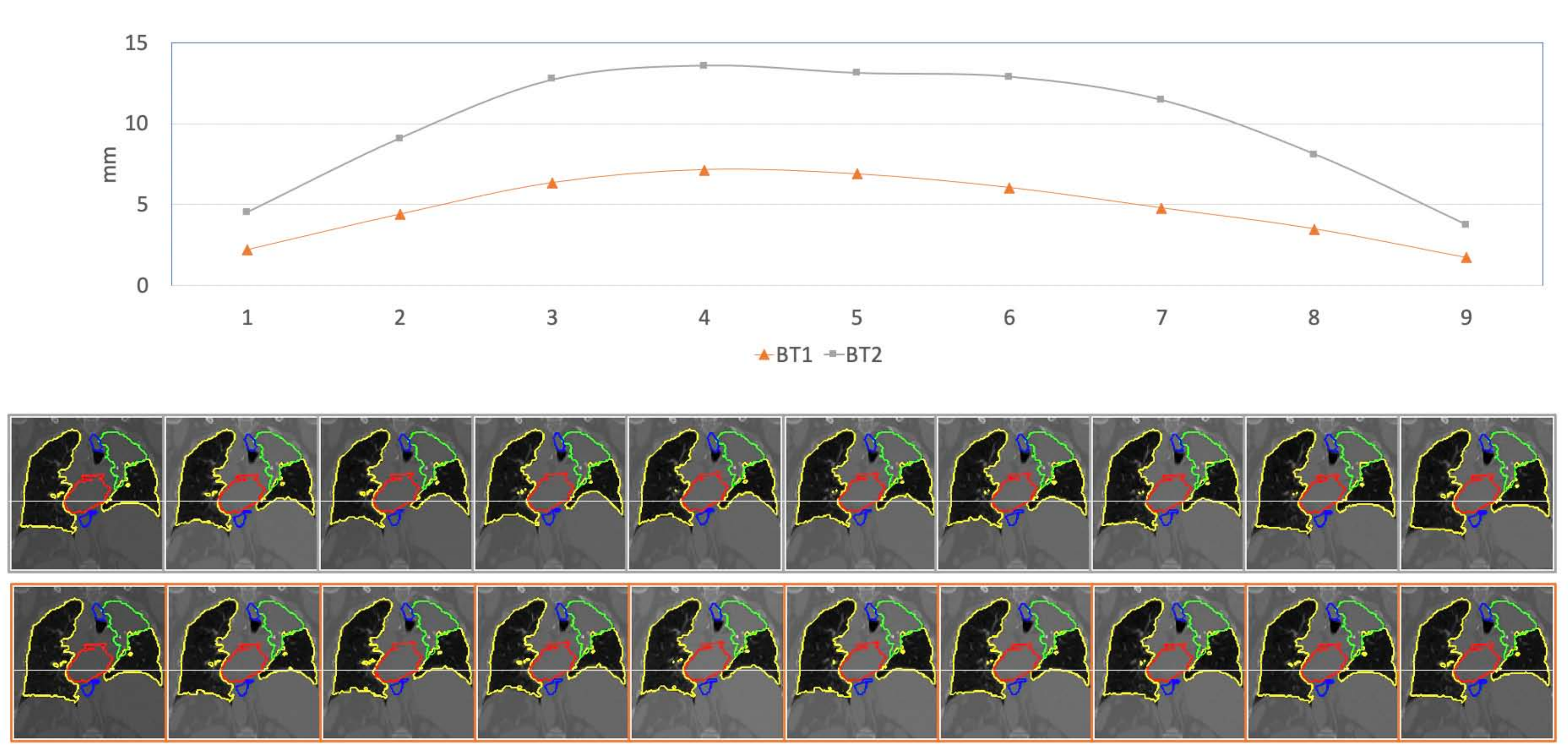}
\caption{Two different breathing traces, BT1 and BT2 shown in the plot, were used to simulate the respiration motion of an internal case, resulting in 2 series of modulated phase images according to the breathing traces. The diaphragm has larger displacement in images simulated with BT2 (upper row) than the displacement in images simulated with shallower BT1 (bottom row.)  The white horizontal line indicates the position of the apex of the left diaphragm at the initial phase (left-most column.) We also overlay the propagated lung(in yellow), heart(in red), esophagus(in blue) and tumor(in green) contours using predicted DVFs.}
\label{fig:Results}
\end{center}
\end{figure}
{The amplitude range in our internal dataset was 0.14 -- 40 mm. To validate the prediction performance on out-of-range displacement, we predicted additional sequences using a 5 times larger respiratory amplitude. The prediction results using a 5 times larger respiratory signal achieve a higher diaphragm level which means the predicted respiratory has larger fluctuation than the original respiratory signal but it was not proportional to the respiratory signal that was used for inference (see Figure~\ref{fig:out_of_range}).}

\begin{figure}[th!]
\begin{center}
\footnotesize
\setlength{\tabcolsep}{3pt}
\includegraphics[width=1\textwidth]{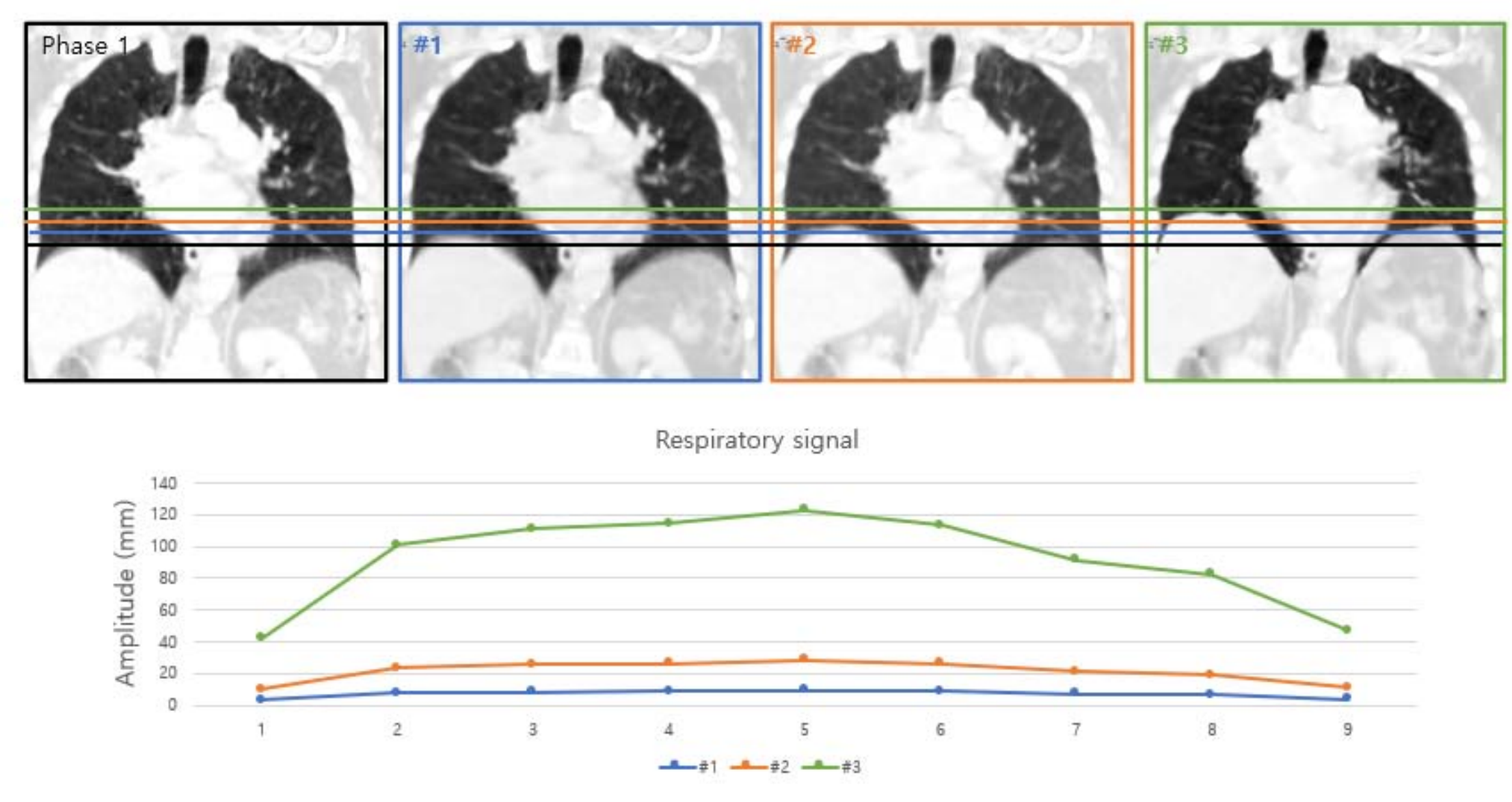}
\caption{The predicted phase 5 images using different 1D respiratory signal. Blue line is original respiratory signal, orange line is 3 times amplitude and green line is 5 times amplitude.}
\label{fig:out_of_range}
\end{center}
\end{figure}

The results of propagating anatomical structures using the predicted DVFs are also shown in Figure~\ref{fig:Results}. We propagated the lung, heart, esophagus, and tumor from the initial phase image. The propagated contours are well-matched with the predicted image and the motion of structures looks very realistic. We also provided the \textbf{supplementary video} of the simulated 4D-CT along with the ground truth 4D-CT and the 3D volume-rendered visualizations. Specifically, the 3D volume-rendered visualizations on LUNA challenge datasets as well as internal lung radiotherapy datasets with structure propagation are included in the accompanying \textbf{supplementary video} with chained predictions for 60-phase predictions for LUNA challenge (radiology lung nodule) and 30-phase predictions for the lung radiotherapy datasets. 

In POPI dataset, there is only one case which contains lung segmentations on all the phases. For this case, we extracted 1D breathing trace from the lung segmentations as we did for our internal dataset. RMSim trained with our internal dataset predicted the remaining phases from the inhale phase with the modulation from the 1D breathing trace. The average TRE (Target Registration Error) of landmarks propagated with our predicted DVFs in this case was 0.92$\pm$0.64mm, showing that RMSim can accurately predict the patient-specific motion from the patient's 1D breathing trace. Figure \ref{fig:POP4DCTTRE} shows the TRE results for all predicted phases in this case. For the three other 4D-CT cases in POPI there were no lung segmentation masks so we performed semi-automatic lung segmentation for extracting the 1D breathing traces and the results are shown in Figure~\ref{fig:dir_valid_supp}.

\begin{figure}[t!]
\begin{center}
\footnotesize
\setlength{\tabcolsep}{3pt}
\includegraphics[width=0.8\textwidth]{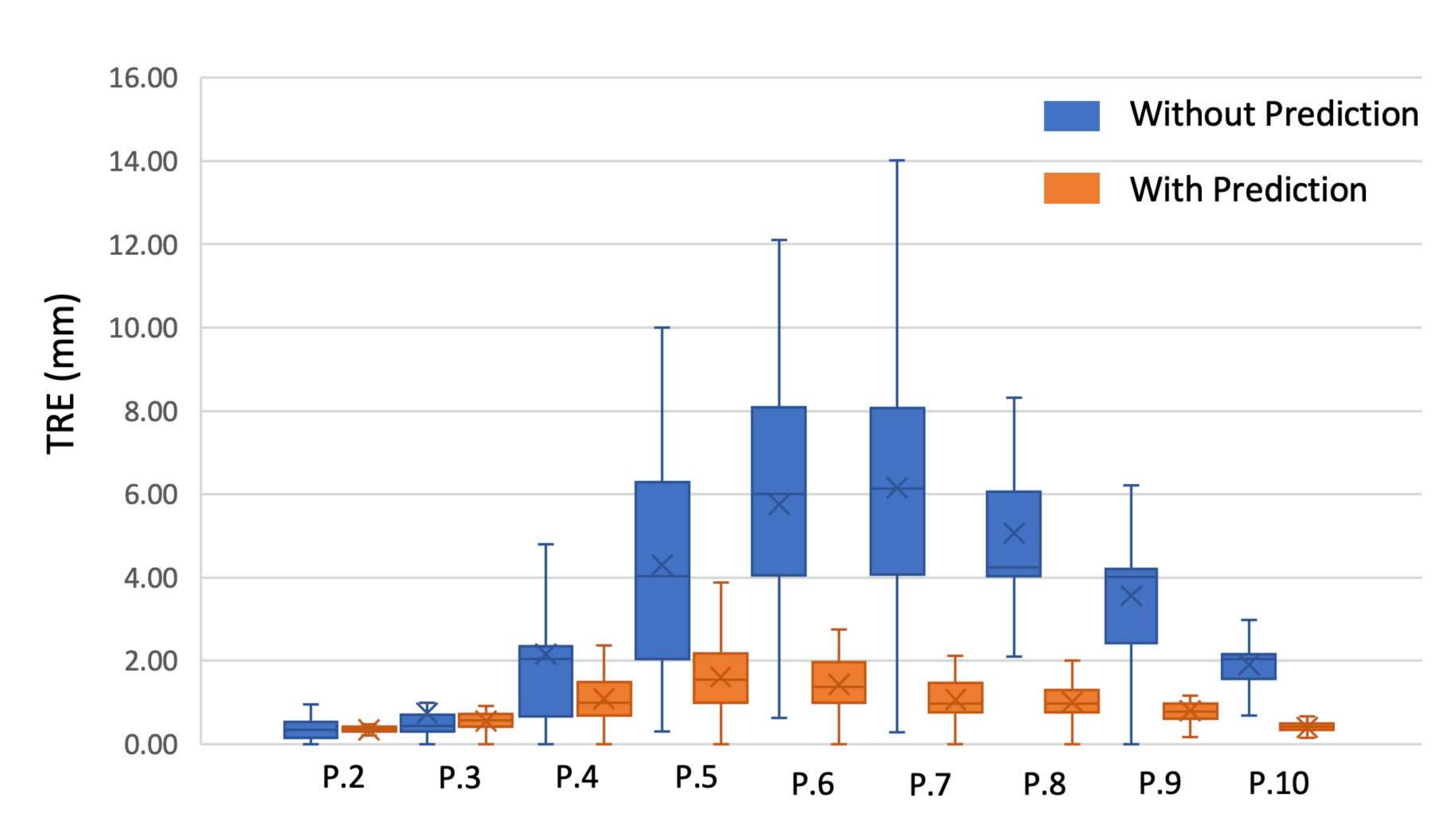}
\caption{TRE results of all 9 phases from the 4DCT case in POPI. RMSim trained with the internal dataset were able to achieve sub-mm accuracy in this external case.}
\label{fig:POP4DCTTRE}
\end{center}
\end{figure}

\begin{figure}[!ht]
\begin{center}
\footnotesize
\setlength{\tabcolsep}{3pt}
\includegraphics[width=1\textwidth]{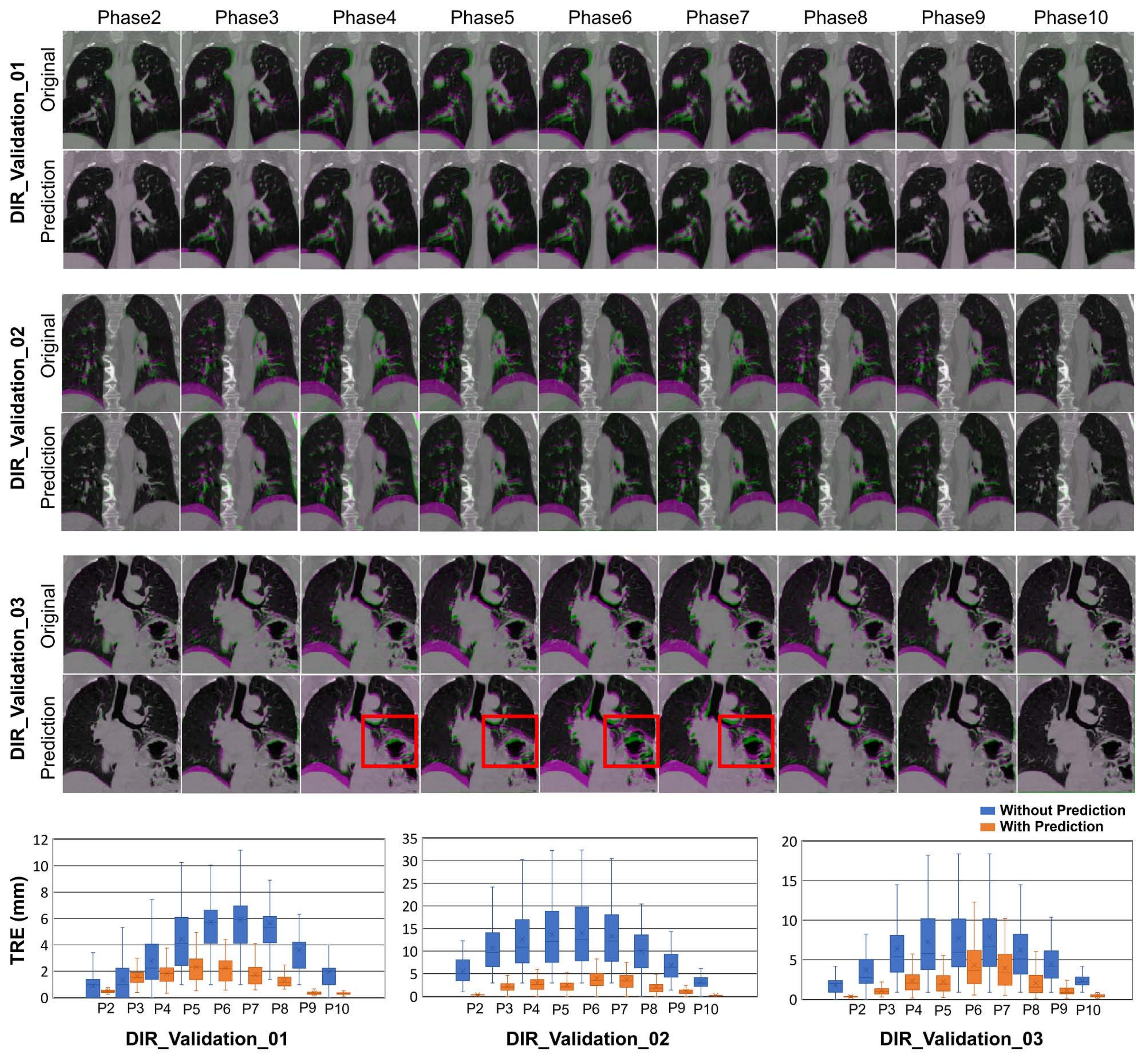}
\caption{Three other 4D-CT POPI cases including 10 phases with landmarks on each phase (TRE plots for the three cases given below). For each case, we show original and predicted phase images overlaid with the difference with respect to original phase 1 input. In original DIR\_Validation\_03 phase difference image, the diaphragm in the left lung (viewer's right) did not move due to the large tumor but it does in our prediction (shown in red bounding boxes). This case does not deflect from the goals of this paper, i.e. data augmentation and DIR validation. The difference in Case \#1 appears minor because the breathing is shallower (less diaphragm movement) and Case \#2 and Case \#3 have larger differences due to deeper breathing.}
\label{fig:dir_valid_supp}
\end{center}
\end{figure}

Additionally, we used the RMSim for augmenting the Learn2Reg Challenge dataset. The Dice score of lung segmentation of 10 Learn2Reg testing cases using the {VoxelMorph without augmentation} was $0.96$ $\pm$ $0.01$ while the model trained with RMSim data augmentation was $0.97$ $\pm$ $0.01$ ($p$ $<$ 0.001 using the \textit{paired t-test}). The SSIM between the warped images and the ground truth images was $0.88$ $\pm$ $0.02$ for the {model without augmentation} and $0.89$ $\pm$ $0.02$  ($p$ $<$ 0.001) for the model with augmentation. 

To validate the improvement of DIR using VoxelMorph with augmentation, we propagated the landmark points from the inhale phase to the exhale phase for the 6 cases available in POPI dataset and computed the TRE. On average, pre-DIR TRE was 8.05$\pm$5.61mm, {VoxelMorph w/o augmentation} was 8.12$\pm$5.78mm compared to 6.58$\pm$6.38mm for VoxelMorph with augmentation ($p$ $<$ 3e-48). The TRE comparison of all 6 cases are shown in Figure \ref{fig:POPI_bar}.

\begin{figure}[t!]
\begin{center}
\footnotesize
\setlength{\tabcolsep}{3pt}
\includegraphics[width=1\textwidth]{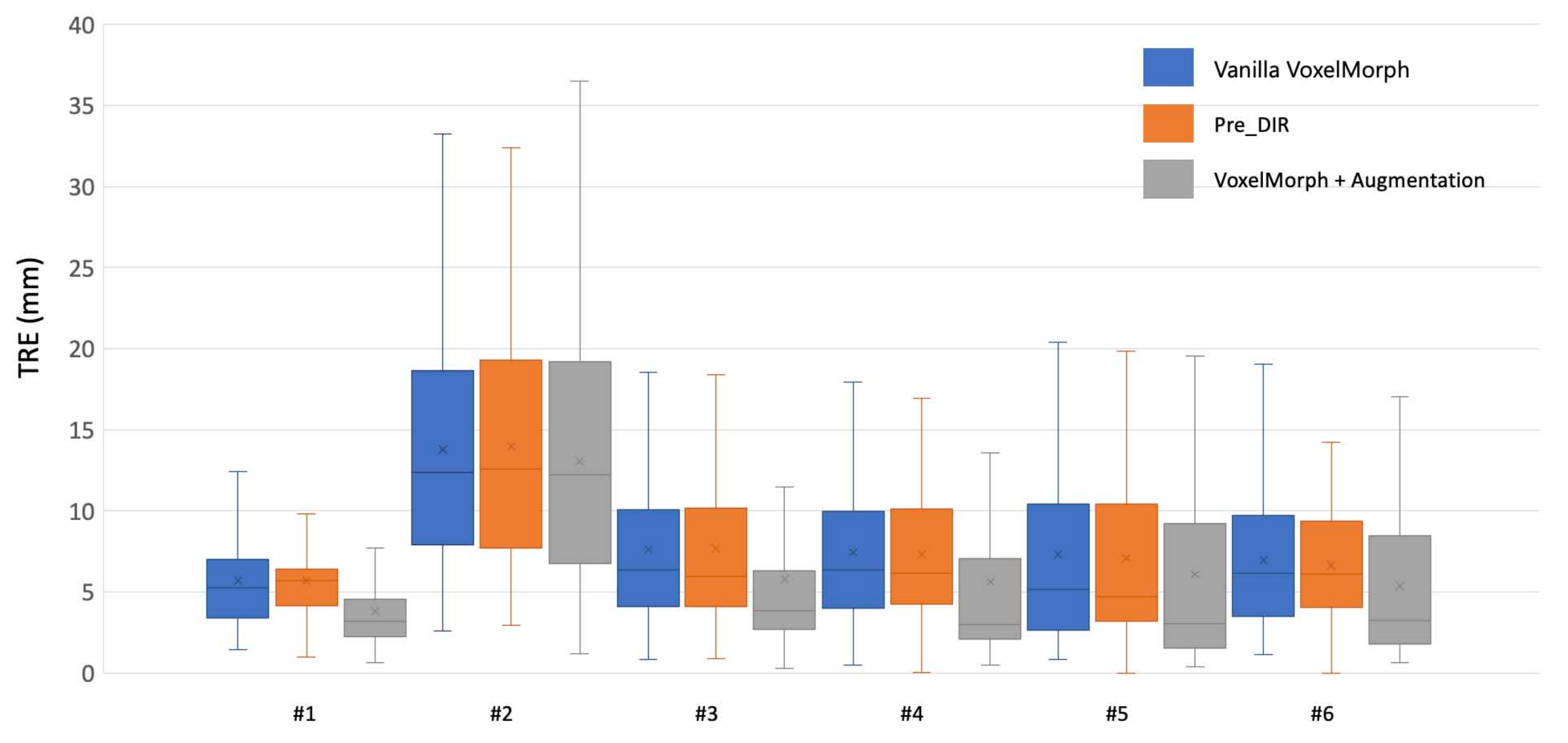}
\caption{TRE results of POPI dataset. VoxelMorph with RMSim augmentation outperformed the {VoxelMorph w/o augmentation} in all 6 cases.}
\label{fig:POPI_bar}
\end{center}
\end{figure}

\section{Discussion}
  



In this work, we presented a 3D Seq2Seq network, referred to as RMSim, to predict patient-specific realistic motion induced/modulated with 1D breathing trace. We successfully validated our RMSim output with both {private and public benchmark datasets (healthy and cancer patients) and demonstrated that adding our patient-specific augmentations to training data can improve performance/accuracy of state-of-the-art deep learning DIR algorithms. We also showcased breathing trace-modulated respiratory motion simulations for public static radiology scans.} In this work, we predicted the motion in one breathing cycle. In the future, we will fine-tune our current model to predict multiple cycles in one-shot. Possible solutions include making our model bi-directional and using cross-attention to improve temporal dynamics in a long sequence. Further research is needed to investigate the impact of training data augmentation on different image modalities such as 4D-MRI.

Another application of our work is in external radiotherapy treatment planning. RMSim simulated 4D-CT can be used to delineate the internal target volume (ITV) which is the union of the target volumes in all respiratory phases. The entire ITV is irradiated in radiation therapy to ensure all regions of tumor receive enough radiation. There is a more sophisticated alternative to ITV, referred to as robust treatment planning, where the key idea is to model the motion and directly incorporate it into the planning \cite{unkelbach_robust_2018}. This typically can be done by assuming a probability density function (PDF) for the position of the target and doing plan optimization based on that~\cite{lens2017probabilistic,watkins2014multiple}. It is also possible to assume a set of possible motion PDFs to account for uncertainty in breathing and plan accordingly \cite{heath2009incorporating, bortfeld2008robust}. The simulated 4D-CT can be used to extract the motion PDF or a set of motion PDFs from varied breathing patterns exhibited by the patient. 

{Additional interesting future direction is the extension of our earlier work in exhaustively simulating physics-based artifacts in CT and CBCT images for more robust cross-modal deep learning translation, segmentation, and motion-correction algorithms \cite{alam2021generalizable,alam2021motion,dahiya2021multitask},
available via our
Physics-ArX 
library 
(\url{https://github.com/nadeemlab/Physics-ArX}).
Specifically, in our previous work we presented a proof-of-concept pipeline for physics-based motion artifact simulation in CT/CBCT images using 4D-CT phases \cite{alam2021motion}. 
Using the method proposed in
the current paper, we can generate and modulate large/diverse 4D-CT phases from any static
3D CT scan using the 1D RPM signal. These generated 4D-CT variations can then be used to
produce large realistic motion-artifact variations via our earlier pipeline}\cite{alam2021motion}.}

\noindent

\textbf{Limitations:} For simplicity, we used the maximal displacement on the diaphragm as the surrogate of clinical breathing trace to drive the modulation. We assume (1) the breathing pattern is regular since we extracted the diaphragm displacements from amplitude-binned 4D-CT, and (2) regional DVFs are linearly scaled according to diaphragm motion. Note 1D breathing trace might not represent the actual cardiac motion. Because of the GPU memory constraints, our input and output dimension was limited to 128x128x128. Nevertheless, the precise estimation of motion is not required for providing realistic motion-induced ground truth DVFs for the validation of the DIR algorithms and data augmentation for training DIR algorithms, as shown in this work. To extend our work to tumor tracking during radiation treatment, we will use the signals from the actual external real-time motion management (RPM) device to drive the modulation more precisely. We will also explore incorporating 2D MV/kV projections acquired during the treatment to infer more realistic cardiac/tumor motion.

\section*{Acknowledgements}
This work was supported partially by NCI/NIH P30 CA008748.


\section*{Conflict of interest}
We have no conflict of interest to declare.

\section*{Code Availability Statement}
The code, pretrained models, and augmented DIR validation datasets will be released at
\url{https://github.com/nadeemlab/SeqX2Y}.

\section*{Data Availability Statement}
The public datasets used in this study and their urls are as follows: (1) Learn2Reg Challenge Lung CT dataset (Empire10 Challenge Dataset): \url{https://drive.google.com/drive/folders/1yHWLQEK9c1xzggkCC4VX0X4To7BBDqu5}, (2) LUNA challenge dataset (subset0.zip): \url{https://zenodo.org/record/3723295}, (3) DIR Validation POPI Dataset (6 4D CT patients with landmarks): \url{https://www.creatis.insa-lyon.fr/rio/dir_validation_data}, and (4) POPI model dataset (one 4D CT patient dataset with landmarks on all phases as well as lung segmentation mask): \url{https://www.creatis.insa-lyon.fr/rio/popi-model_original_page}. 



\section*{References}

\end{document}